\documentclass[11pt]{article}

\usepackage[a4paper,margin=1in]{geometry}

\usepackage[T1]{fontenc}
\usepackage[utf8]{inputenc}
\usepackage{lmodern}

\usepackage{amsmath,amssymb}

\usepackage{graphicx}
\usepackage{xcolor} %

\usepackage{booktabs}

\usepackage{hyperref}

\usepackage{float}

\usepackage[numbers]{natbib}

\usepackage{calligra}

\newcommand{\ejw}[1]{\textcolor{purple}{\kern-1pt\textsc{\textbf{\calligra{E.J.W.}}}\kern-1pt\textit{\Large\calligra{}}\kern-1pt: \textit{#1}}}

\usepackage{amsmath,amsfonts,bm}

\def\eqref#1{(\ref{#1})}

\def\1{\bm{1}}

\DeclareMathAlphabet{\mathsfit}{\encodingdefault}{\sfdefault}{m}{sl}
\SetMathAlphabet{\mathsfit}{bold}{\encodingdefault}{\sfdefault}{bx}{n}

\def\x{{\mathbf x}}

\def\z{{\mathbf z}}

\definecolor{ourblue}{rgb}{0.368,0.507,0.71}
\definecolor{ourorange}{rgb}{0.881,0.611,0.142}
\definecolor{ourgreen}{rgb}{0.56,0.692,0.195}
\definecolor{ourred}{rgb}{0.923,0.386,0.209}
\definecolor{ourviolet}{rgb}{0.528,0.471,0.701}
\definecolor{ourbrown}{rgb}{0.772,0.432,0.102}
\definecolor{ourlightblue}{rgb}{0.364,0.619,0.782}
\definecolor{ourdarkgreen}{rgb}{0.572,0.586,0.}
\definecolor{url}{HTML}{d95225}
\definecolor{bloodred}{HTML}{B00000}

\definecolor{ourcyan2}{rgb}{0.125,0.722,0.804}
\definecolor{ourred2}{rgb}{0.863,0.184,0.047}
\definecolor{ouryellow2}{cmyk}{0,0.16,1.0,0.07}
\definecolor{ourviolet2}{cmyk}{0.55,0.56,0,0.47}
\definecolor{ourorange2}{cmyk}{0,0.46,0.89,0.11}

\title{\vspace{-1cm}\Large \bfseries Mercury: Ultra-Fast Language Models Based on Diffusion}
\author{
    \begin{tabular}{c}
        Inception Labs\\
        Samar Khanna*, Siddhant Kharbanda*, Shufan Li*, Harshit Varma*, Eric Wang*\\
Sawyer Birnbaum$^\wedge$, Ziyang Luo$^\wedge$, Yanis Miraoui$^\wedge$, Akash Palrecha$^\wedge$\\
Stefano Ermon$^\sharp$, Aditya Grover$^\sharp$, Volodymyr Kuleshov$^\sharp$\\
*$^\wedge$$^\sharp$ equal core, cross-function, senior contributors listed alphabetically.
        \\
        \texttt{hello@inceptionlabs.ai}
    \end{tabular}
}
\date{\vspace{-2ex}}

\begin{document}
\maketitle

\begin{center}
    \textbf{Abstract}
\end{center}
\noindent

We present Mercury, a new generation of commercial-scale large language models (LLMs) based on diffusion.
These models are parameterized via the Transformer architecture and trained to predict multiple tokens in parallel.
In this report, we detail Mercury Coder, our first set of diffusion LLMs designed for coding applications.
Currently, Mercury Coder comes in two sizes: Mini and Small.
These models set a new state-of-the-art on the speed-quality frontier.
Based on independent evaluations conducted by Artificial Analysis, Mercury Coder Mini and Mercury Coder Small achieve state-of-the-art throughputs of 1109 tokens/sec and 737 tokens/sec, respectively, on NVIDIA H100 GPUs and outperform speed-optimized frontier models by up to $10\times$ on average while maintaining comparable quality.
We discuss additional results on a variety of code benchmarks spanning multiple languages and use-cases as well as real-world validation by developers on Copilot Arena, where the model currently ranks second on quality and is the fastest model overall. 
We also release a public API at \url{platform.inceptionlabs.ai} and free playground at \url{chat.inceptionlabs.ai}. 

\vspace{1em}

{\hypersetup{linkcolor=black}
\tableofcontents
}
\vspace{1em}
\clearpage

\section{Introduction}
\label{sec:intro}

Diffusion models have emerged as the state-of-the-art approach for generating images~\cite{rombach2022high} and videos~\cite{brooks2024video}, consistently producing high-quality, coherent, and diverse content~\cite{sohl2015deep,song2019generative,ho2020denoising}. However, the application of diffusion to discrete data---particularly language---has remained limited to small-scale experiments~\cite{austin2021structured,gulrajani2024plaid,li2022diffusion,lou2023discrete,sahoo2024simple,israel2025enabling}. 
The advantage of diffusion relative to classical autoregressive models lies in its ability to perform parallel generation, which can greatly improve speed, in addition to fine-grained control, reasoning, and multi-modal data processing capabilities. Scaling diffusion models to the size of modern large language models (LLMs)~\cite{achiam2023gpt,team2023gemini,dubey2024llama} while maintaining high performance has remained an open challenge.

In this report, we introduce Mercury---the first family of large-scale diffusion-based language models by Inception Labs. Mercury models achieve state-of-the-art performance and efficiency relative to comparable autoregressive (AR) models. 
Specifically, we present Mercury Coder, a set of Mercury models optimized for code. 
A predominant use-case of generative AI is for coding applications. Over 84\% of developers have experience with code LLMs, highlighting the growing role of generative AI in streamlining software development~\cite{9cv9_2024}. 
However, high per-user latency of prominent use-cases, such as auto-completion, code editing, and agentic workloads, limits wider adoption of coding applications. Accordingly, we focus our first set of Mercury models on coding.

Mercury Coder models demonstrate strong performance on key coding benchmarks, highlighting improved accuracy, correctness, and in-filling capabilities across commonly used programming languages.
By generating tokens in parallel in a coarse-to-fine manner, our models make significantly better use of modern GPU architectures, which leads to a higher arithmetic intensity of the generation algorithm and overall improved computational efficiency. 
This drastically improves user experience, especially for latency-sensitive, decode-heavy applications such as coding assistants, agentic workloads, chain-of-thought reasoning, and edge computing.  As AI inference demand continues to scale, diffusion models can reduce inference costs significantly, making them a more sustainable solution for large-scale AI deployment. 

\begin{figure}[ht]
    \centering
    \includegraphics[width=\linewidth]{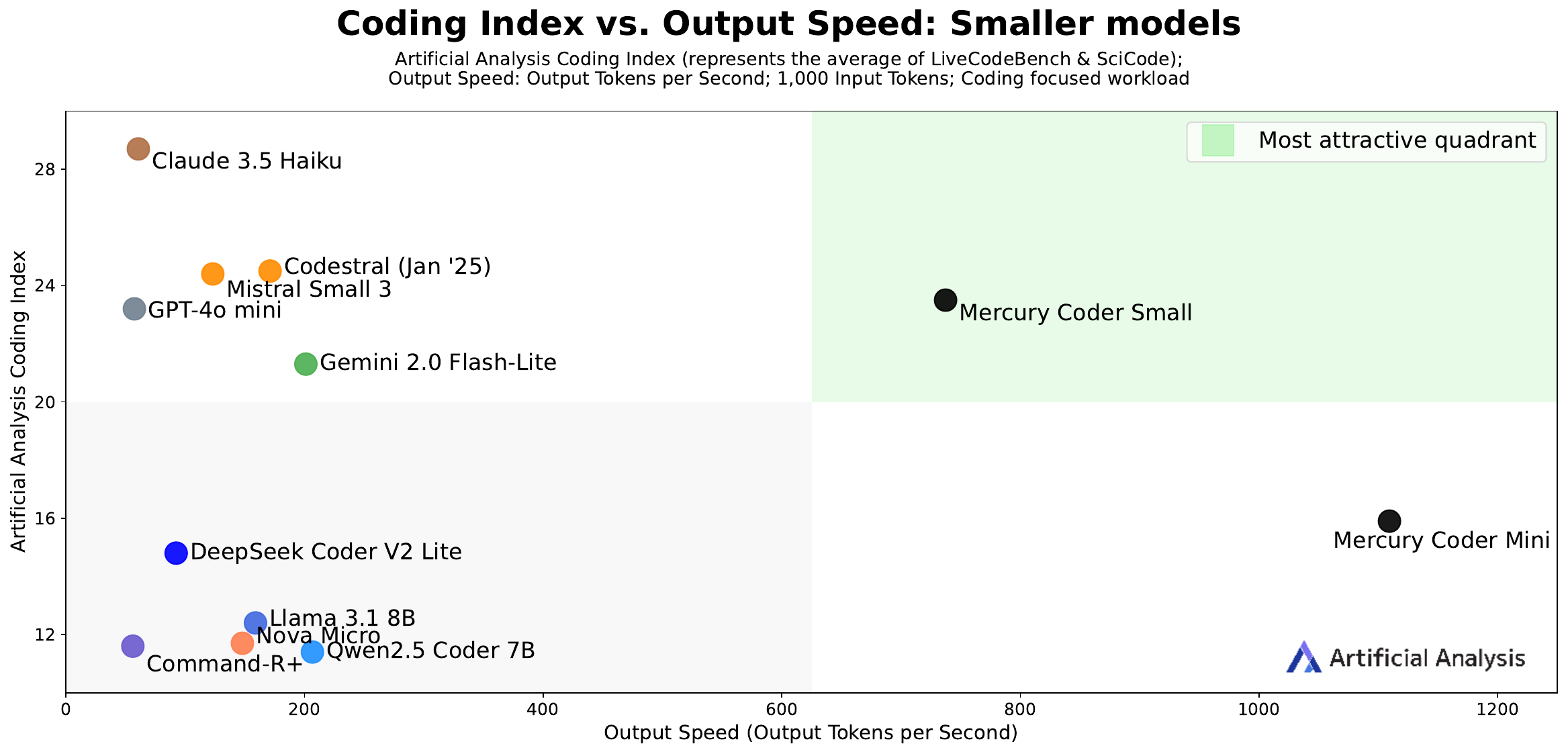}
    \caption{Quality vs. Speed Trade-offs for Mercury Coder models. We find that Mercury Coder models outperform other frontier models by up to 10x in throughput while maintaining comparable quality on challenging code generation benchmarks. Figure taken from third-party evaluations conducted by Artificial Analysis.}
    \label{fig:enter-label}
\end{figure}

Notably, the Mercury models retain a Transformer-based architecture~\cite{vaswani2017attention}, ensuring compatibility with many of the modeling and system-level optimizations developed in recent years for scalable training and inference of large language models.
When prompted with a query, instead of producing the answer one token at a time, the answer is generated in a coarse-to-fine way. Improvements are suggested by a neural network---in our case a Transformer model---which is trained on large amounts of data to globally improve the quality of the answer by modifying multiple tokens in parallel. 
Our models can be easily adapted for diverse applications by leveraging established methodologies for instruction tuning and alignment and can serve as a drop-in replacement for autoregressive models with greatly improved inference-time efficiency.

In the following sections, we detail the architecture, performance metrics, and potential applications of our diffusion-based language models. Our work represents a step toward more efficient, scalable, and controllable AI systems, with broad implications for the future of text generation and multi-modal AI.

\subsection{Contributions}
\begin{itemize}
\item This paper describes the Mercury family of diffusion large language models (dLLMs), a new generation of LLMs that push the frontier of fast, high-quality text generation.

\item Mercury is up to 10x faster than frontier speed-optimized LLMs. Our models run at over 1000 tokens/sec on NVIDIA H100s, a speed previously possible only using custom chips.

\item In addition to ultra-fast speeds, our coding models are comparable in quality to high-speed commercial offerings on coding benchmarks covering diverse usecases, programming languages, and  hardware backends.
\end{itemize}

\section{Inception Mercury Model Family}
\label{sec:inception-models}

\label{subsec:inception-family}

This report introduces the Inception Mercury models, a line of speed-optimized dLLMs.
Our first focus is on coding models. Coding is a highly latency sensitive domain and the ability to generate fast code directly influences user experience, agentic workloads, and complex reasoning. We present two models in the Mercury Coder series.

\begin{enumerate}

    \item \textbf{Mercury Coder Mini} Our Mini model features the highest speed as well as competitive quality. For the first time, we attain throughputs of 1100+ tokens/second on H100 GPUs in latency-optimized regimes, while maintaining quality comparable to that of popular speed-optimized, open-weights models. 
    \item \textbf{Mercury Coder Small} Our Small model achieves benchmark performance that matches popular speed-optimized frontier models, while having 3-10x better throughput in latency-optimized regimes. They achieve speeds of 700+ tokens/second across coding workloads.
\end{enumerate}

\subsection{Training}

The Mercury diffusion models are defined by a generation process that iteratively refines outputs in parallel starting from random noise and gradually transforming it into a sample from the data distribution.
Our methods extend \cite{lou2023discrete} through careful modifications to the data and computation to scale up learning.
The overall model is trained on the order of trillions of tokens.
The training data comprises a combination of web crawls along with carefully curated real and synthetic datasets derived from proprietary data sources.
We conduct all our development on a large-scale cluster of NVIDIA H100s.

More formally, we define our diffusion models via a pair of forward and reverse processes. The forward or noising process $q$ starts from clean data $\x \in \mathcal{X}$ (a sequence of natural language tokens, e.g., a sequence of words) and defines a set of latent variables $\z_t \in \mathcal{X}$ over time steps $t=1,...,T$ via a Markov process denoted as $q(\z_t|\z_{t-1})$. 
The latents $\z_t$ represent increasingly noisy versions of $\x$, and the final $\z_T$ are designed to be distributed according to a known prior noise distribution $p(\z_T)$. 
The reverse or denoising process $p$ generates data
by first sampling $\z_T \sim p(\z_T)$ and 
then applying a model $p(\z_{t-1}|\z_t)$ to iteratively denoise the data.
This procedure defines a probability distribution $p(\x)$.

The model $p$ is defined by learned parameters $\theta$, hence we denote it by $p_\theta$.
The parameters are chosen to minimize a loss that fits $p$ to reverse $q$.
In practice, this can be achieved by first learning a denoising model, i.e., by minimizing
\begin{align*}\label{eqn:elbo}
    \mathcal{L}(\x) = - \mathbb{E}_t \left[ \gamma(t) \cdot \mathbb{E}_{\z_t \sim q} \log p_\theta( \x | \z_t) \right],
\end{align*}
where $\gamma(t) \geq 0$ is a user-specified function that assigns a weight to each noise level and $p_\theta( \x | \z_t)$ is a distribution over clean data $\x$ given noisy data $\z_t$.
The denoiser can then be used for generation e.g., by defining $p_\theta(\z_{t-1}|\z_t) = \sum_{x} q(\z_{t-1}|\z_t,\x)p_\theta(\x|\z_t)$.

\paragraph{Architecture}
\label{subsec:architecture}

Inception Mercury models are based on a Transformer architecture~\cite{vaswani2017attention}. 
Note that this choice of architecture is orthogonal to the fact that the Mercury models are diffusion-based. Diffusion implies specific training and generation algorithms, but does not pose constraints on the architecture of neural network that is trained. For example, a dLLM could also be based on a recurrent architecture~\cite{peng2023rwkv,gu2023mamba}. This is analogous to architecture choices for image diffusion models, in which the denoising network can also be parameterized with a U-Net~\cite{ho2020denoising} or a transformer~\cite{peebles2023scalable}.
Relying on a Transformer architecture has a number of advantages. It allows Mercury models to benefit from efficient implementations of low-level primitives, and it simplifies hyper-parameter search and optimization.

\paragraph{Fine-tuning and Alignment}
Inception Mercury Models can benefit from further pre-training, fine-tuning and alignment on downstream datasets via RLHF~\cite{ouyang2022training} or DPO~\cite{rafailov2023direct} techniques to improve downstream performance. 
The key change for all stages is to replace the autoregressive loss with a denoising diffusion loss.

\paragraph{Context Length}
Inception Mercury models support a context length of up to 32,768 tokens out of the box and up to 128k tokens with context extension approaches. 
This protocol follows standard training recipes used for developing language models~\cite{dubey2024llama,yang2024qwen2,liu2024deepseek}.

\subsection{Inference}
\paragraph{Prompting}

In addition to generating full sequences from scratch, our inference methods support flexible generation conditioned on a prompt or context.
Given that the Mercury models support conditional generation, and given that they can be trained, fine-tuned, and aligned on datasets that are analogous to those of traditional language models, the Mercury models also support prompting as in traditional LLMs. This includes zero-shot prompting, few-shot prompting~\cite{brown2020language}, and chain-of-thought~\cite{wei2022chain}.

\paragraph{Serving}

While prior diffusion models such as \cite{lou2023discrete} show that it is possible to reduce the number of forward pass iterations for sub-billion parameter models, they fail to show improvements in wall-clock efficiency.
From a systems perspective, our algorithm’s speed advantages owe to its maximum utilization of the computing power available on commonly available hardware accelerators, such as NVIDIA GPUs. 
To ensure maximum speed, we rely on a proprietary inference engine that implements highly efficient diffusion sampling. 
The engine features a dynamically batched sampling and paging implementation that can automatically navigate the speed/quality trade-off under production workloads.
To push performance even further, we leverage a set of custom kernels for parallel inference workloads.
From a user's perspective, we can expose to the user an API compatible with the OpenAI standard.  This backwards compatiblity with existing APIs enables Mercury to serve as a drop-in replacement for autoregressive models.

\section{Capabilities}
\label{sec:capabilities}

This section provides an in-depth analysis on the capabilities of Mercury
with regards to quality and decoding efficiency. %
Our model was tested on an API endpoint hosted in February 2025. 

\subsection{Baselines}

We benchmark Mercury against four sets of autoregressive LLM baselines. These sets of models target different use cases and strike a different balance of accuracy and speed.

\paragraph{Open-Weights Speed-Optimized Models} 

We compare against models from the Llama 3.1~\cite{Dubey2024TheL3}, Qwen 2.5~\cite{Hui2024Qwen25CoderTR}, Mistral~\cite{mistral_small_3}, and DeepSeek V2~\cite{DeepSeekAI2024DeepSeekCoderV2BT} families.

\paragraph{Open-Weights Frontier Models}  
In this category, we compare against DeepSeek V3~\cite{DeepSeekAI2024DeepSeekV3TR} which is comparable to Claude 3.5 Sonnet~\cite{TheC3} and GPT 4o~\cite{Hurst2024GPT4oSC} in performance, while being open-weights.

\paragraph{Closed-Weights Speed-Optimized Models}  
These proprietary models provide low per-token costs and fast inference speeds, often targeting deployment in latency-sensitive environments and simpler tasks. They strike a balance between performance and cost, and can match frontier performance on tasks like summarization and auto-completion. For our evaluations, we consider models from the Claude 3.5~\cite{TheC3}, GPT 4o~\cite{Hurst2024GPT4oSC}, Gemini 2.0 Flash~\cite{deepmind_gemini_flash_2_0}, Amazon Nova~\cite{Intelligence2024}, and Codestral~\cite{mistral2025codestral} families. 

\paragraph{Closed-Weights Frontier Models}  
Closed-weights frontier models represent the state-of-the-art in language model performance. These models are typically at the top of LLM benchmarks; however, they are typically not publicly accessible for alignment or fine-tuning. In our comparisons, we include the leading proprietary models (GPT 4o~\cite{Hurst2024GPT4oSC}, Claude 3.5 Sonnet~\cite{TheC3}). Note however, that Mercury models are in a speed-optimized class that targets a different speed-cost-performance trade-off from frontier models; we include these numbers only for context.

\begin{table}[t]
    \centering
    
    \caption{Performance (pass@1) comparison of various models across different coding benchmarks, grouped by model category. `\textbf{*}' indicates metrics as reported by Artificial Analysis.}
    \label{tab:coding_benchmarks}

\resizebox{\textwidth}{!}{
\begin{tabular}{lccccccc}
        \toprule
        \textbf{Model} & \textbf{HumanEval*} & \textbf{MBPP} & \textbf{EvalPlus} & \textbf{MultiPL-E} & \textbf{LCB*} & \textbf{BCB} & \textbf{Speed} \\
        \midrule
        \multicolumn{8}{l}{\textbf{Open-Weights Models}} \\
        Llama 3.1 8B Instruct & 66.5 & 59.2 & 60.2 & 50.1 & 12.0 & 32.3 & 153 \\
        DeepSeek Coder V2 Lite & 79.0 & 59.8 & 68.3 & 57.0 & 16.0 & 44.4 & 93 \\
        Mistral Small 3 & 84.8 & 69.6 & 72.3 & 70.1 & 25.0 & 42.7 & 126 \\
        Qwen 2.5 Coder 7B Instruct & 88.0 & 80.0 & 79.3 & 75.3 & 9.0 & 41.4 & 195 \\
        \midrule
        \multicolumn{8}{l}{\textbf{Frontier Speed-Optimized Models}} \\
        Nova Micro & 79.3 & 65.4 & 72.1 & 56.7 & 14.0 & - & 148 \\
        Codestral 2501 & 85.0 & 72.2 & 75.6 & 73.4 & 24.0 & 46.1 & 171 \\
        GPT 4o Mini & 88.0 & 74.6 & 78.5 & 72.0 & 23.0 & 46.8 & 59 \\
        Claude 3.5 Haiku & 86.0 & 78.0 & 75.1 & 72.3 & 31.0 & 45.4 & 61 \\
        Gemini 2.0 Flash Lite & 90.0 & 75.0 & 77.3 & 79.5 & 18.0 & 44.4 & 201 \\
        \midrule
        \multicolumn{8}{l}{\textbf{Frontier Models}} \\
        DeepSeek V3 & 92.1 & 81.0 & 82.1 & 79.1 & 36.0 & 50.0 & 27 \\
        Claude 3.5 Sonnet & 90.2 & 81.2 & 77.3 & 81.9 & 38.0 & 44.8 & 76 \\
        GPT 4o & 90.2 & 82.2 & 82.4 & 77.6 & 31.0 & 49.9 & 61 \\
        \midrule
        \multicolumn{8}{l}{\textbf{Our Models}} \\
        Mercury Coder Mini & 88.0 & 77.1 & 78.6 & 74.1 & 17.0 & 42.0 & 1109 \\
        Mercury Coder Small & 90.0 & 76.6 & 80.4 & 76.2 & 25.0 & 45.5 & 737 \\
        \bottomrule
\end{tabular}
}

\end{table}
\subsection{Coding Capabilities}
\label{subsec:capabilities-code}

\subsubsection{Evaluation Benchmarks}

We report the quality of our coding models across standard benchmarks.
HumanEval~\cite{Chen2021EvaluatingLL} and MBPP~\cite{Austin2021ProgramSW} assess Python code generation based on test pass rates. EvalPlus~\cite{Liu2023IsYC} extends the evaluation to more test cases. LiveCodeBench~\cite{Jain2024LiveCodeBenchHA} focuses on more sophisticated coding scenarios. MultiPL-E~\cite{Cassano2022MultiPLEAS} evaluates multi-language code generation across C++, JavaScript, Java, PHP, Bash, and TypeScript. FIM~\cite{Bavarian2022EfficientTO} measures code in-filling ability targeting autocomplete-like scenarios.

\paragraph{Speed}
We compare the speed of Mercury to that of existing autoregressive models. We evaluate the speed of an end-to-end deployment of each type of model (i.e., we compare APIs). 
In the context of Mercury models, we evaluate a deployment on our custom serving engine on Nvidia hardware.

We report results from an independent third-party evaluation of various APIs by the firm Artificial Analysis (AA)\footnote{\url{https://artificialanalysis.ai/}}. The evaluation relies of a series of coding-focused prompts featuring approximately 1,000 input and 1,000 output tokens and that are proprietary to AA.
We use throughput (measured in output tokens/second) as our main measure of speed.
Throughput is measured by performing inference on a target dataset and dividing the processing time 
from the first to that output token by the number of output tokens in the dataset. 

In order to compare the end-to-end speed of Mercury to existing models, we report the speed of commercial APIs for these models, as estimated by Artificial Analysis. These speed measurements correspond to a median throughput benchmarked by AA across cloud providers serving the model.

\subsubsection{Results}

\begin{table}[t]
\centering
\caption{Performance comparison of various models on the MultiPL-E benchmark across different programming languages (values in \%).}
\begin{tabular}{lccccccc}
\toprule
\textbf{Model} & \textbf{CPP} & \textbf{Java} & \textbf{JS} & \textbf{PHP} & \textbf{Bash} & \textbf{TS} & \textbf{Avg} \\
\midrule
\multicolumn{8}{l}{\textbf{Open-Weights Models}} \\
Llama 3.1 8B Instruct & 54.0 & 48.7 & 57.8 & 49.1 & 34.8 & 55.9 & 50.1 \\
OpenCoder 8B Instruct & 70.2 & 70.8 & 78.9 & 72.1 & 44.1 & 75.2 & 68.5 \\
Mistral Small 3 & 74.5 & 73.9 & 82.0 & 66.5 & 43.5 & 80.1 & 70.1 \\
Qwen 2.5 Coder 14B Instruct & 77.6 & 55.9 & 83.9 & 61.5 & 46.0 & 83.9 & 68.1 \\
\midrule
\multicolumn{8}{l}{\textbf{Frontier Speed-Optimized Models}} \\
Nova Micro & 59.6 & 57.1 & 67.7 & 62.1 & 23.6 & 70.2 & 56.7 \\
Codestral 2501 & 80.1 & 72.7 & 83.2 & 73.9 & 47.2 & 83.2 & 73.4 \\
GPT 4o Mini & 78.3 & 73.4 & 82.0 & 71.4 & 46.6 & 80.1 & 72.0 \\
Claude 3.5 Haiku & 75.2 & 78.5 & 79.5 & 73.3 & 45.3 & 82.0 & 72.3 \\
Gemini 2.0 Flash Lite & 84.5 & 82.6 & 88.2 & 85.7 & 50.9 & 85.1 & 79.5 \\
\midrule
\multicolumn{8}{l}{\textbf{Frontier Models}} \\
DeepSeek V3 & 84.5 & 73.9 & 87.6 & 81.4 & 59.0 & 88.2 & 79.1 \\
Claude 3.5 Sonnet & 82.0 & 88.6 & 89.4 & 83.2 & 57.8 & 90.6 & 81.9 \\
GPT 4o & 79.5 & 81.0 & 87.0 & 78.3 & 52.8 & 87.0 & 77.6 \\
\midrule
\multicolumn{8}{l}{\textbf{Our Models}} \\
Mercury Coder Mini & 78.9 & 74.5 & 78.9 & 72.7 & 56.5 & 83.2 & 74.1 \\
Mercury Coder Small & 82.0 & 80.1 & 83.9 & 78.3 & 50.1 & 82.6 & 76.2\\
\bottomrule
\end{tabular}

\label{tab:multipl-e-results}
\end{table}

\begin{table}[h]
    \centering
    \caption{Performance comparison of various models on the fill-in-the-middle (FIM) single-line and random-span-light benchmarks, grouped by model category.}
    \resizebox{\textwidth}{!}{
    \begin{tabular}{lccc}
        \toprule
        \textbf{Model} & \textbf{FIM Single-Line} & \textbf{FIM Random-Span-Light} & \textbf{Average} \\
        \midrule
        \multicolumn{4}{l}{\textbf{Open-Weights Models}} \\
        Llama 3.1 8B Instruct & 37.9 & 11.0 & 24.5 \\
        DeepSeek Coder V2 Instruct Lite & 55.1 & 47.0 & 51.1 \\
        Qwen 2.5 Coder 7B Instruct & 89.6 & 56.1 & 72.9 \\
        \midrule
        \multicolumn{4}{l}{\textbf{Frontier Speed-Optimized Models}} \\
        Claude 3.5 Haiku & 63.6 & 27.4 & 45.5 \\
        Gemini 2.0 Flash Lite & 65.4 & 54.9 & 60.2 \\
        GPT 4o Mini & 74.8 & 47.0 & 60.9 \\
        Codestral 2501 & 93.0 & 72.0 & 82.5 \\
        \midrule
        \multicolumn{4}{l}{\textbf{Our Models}} \\
        Mercury Coder Mini & 92.9 & 71.5 & 82.2 \\
        Mercury Coder Small & 93.1 & 76.5 & 84.8 \\
        \bottomrule
    \end{tabular}
    }
    \label{tab:ai_model_comparison}
    
\end{table}

Table~\ref{tab:coding_benchmarks} compares the performance of various models on key code generation benchmarks, including HumanEval, MBPP, EvalPlus, MultiPL-E, LiveCodeBench, and BigCodeBench. 

\paragraph{Overall Coding Performance}
Mercury Coder Mini, our smaller model, outperforms all open-weight models while being more than $8\times$ faster and achieving speeds of around $1,100$ tokens per second. This makes it a compelling choice for real-world applications that require high efficiency.
Meanwhile, Mercury Coder Small performs on par with frontier speed-optimized models like Claude 3.5 Haiku and Gemini 2.0 Flash, and is also much faster. 
While some speed-optimized models are fast, there remains a trade-off between latency and accuracy---diffusion models significantly push the Pareto frontier.

\paragraph{Performance Across Programming Languages}

We evaluate multiple code generation models on the MultiPL-E benchmark, assessing their performance across six programming languages: C++, Java, JavaScript, PHP, Bash, and TypeScript. Table~\ref{tab:multipl-e-results} presents the accuracy of each model, measured as the percentage of correct solutions generated. Among open-weight models, Mistral Small 3 and OpenCoder 8B Instruct achieve the highest average performance. Frontier speed-optimized models, such as Gemini 2.0 Flash Lite and Codestral 2501, demonstrate strong results, outperforming many open-weight models while maintaining efficiency. Our models, Mercury Coder Mini and Mercury Coder Small, outperform open-weights models and show competitive performance to well-established speed-optimized models, especially in Java and JavaScript. These results highlight the effectiveness of diffusion in multi-language code generation. %

\paragraph{Fill-in-the-Middle}

We evaluate model performance on fill-in-the-middle (FIM) tasks, assessing their ability to generate missing code in single-line and random-span-light settings. Table~\ref{tab:ai_model_comparison} presents results across different model categories. Among open-weight models, Qwen 2.5 Coder 7B Instruct achieves the highest performance. Frontier speed-optimized models show stronger results, with Codestral 2501 leading the category, followed by GPT-4o Mini and Gemini 2.0 Flash Lite, which maintain a balance between accuracy and efficiency. Our models, Mercury Coder Mini and Mercury Coder Small, achieve state-of-the-art performance in FIM tasks, surpassing all evaluated models, including Codestral 2501. These results highlight the effectiveness of our models in code completion scenarios.

\paragraph{Human Evaluation on Copilot Arena}

\begin{table}[h]
\caption{Co-Pilot Arena model comparison by latency, Elo scores, and ranks. Data obtained via Copilot Arena.}
\centering
\begin{tabular}{lcccc}
\hline
\textbf{Model} & \textbf{Latency (seconds)} & \textbf{Latency Rank} & \textbf{Elo Score} & \textbf{Elo Rank} \\
\hline
DeepSeek V2.5 (FIM) & 2.07 & 11 & 1025 & 1 \\
Claude 3.5 Sonnet & 1.46 & 8 & 1003 & 1 \\
\textbf{Mercury Coder Mini} & \textbf{0.25} & \textbf{1} & \textbf{993} & \textbf{2} \\
Codestral & 0.31 & 2 & 992 & 2 \\
Metal Llama 3.1 405B & 1.84 & 10 & 982 & 3 \\
GPT-4o & 0.76 & 5 & 980 & 3 \\
Gemini 1.5 Flash & 0.59 & 3 & 977 & 3 \\
Gemini 1.5 Pro & 1.48 & 9 & 977 & 5 \\
Meta Llama 3.1 70B & 1.17 & 7 & 970 & 5 \\
Qwen 2.5 Coder 32B & 0.91 & 6 & 949 & 12 \\
GPT-4o Mini & 0.84 & 4 & 939 & 12 \\
\hline
\end{tabular}
\label{tab:model_comparison}
\end{table}

We complement our benchmark results with a human evaluation against other models in the setting of code assistants. Specifically, we evaluated our Mercury Coder Mini on Copilot Arena~\cite{chi2025copilot}, a platform in which users are presented with code completions from different models and provide their preference.

On Copilot Arena, Mercury Coder Mini is tied for second place, surpassing the performance of speed-optimized models like GPT-4o Mini and Gemini-1.5-Flash and even of larger models like GPT-4o. At the same time, it is the fastest model, with an average latency of just 25 ms, about 4 times faster than GPT-4o Mini. 

\paragraph{Scaling}

Modern large language models scale in performance as their size and training data increase. While most research focuses on autoregressive models, the scaling properties of diffusion large language models are less well understood. We observe that the performance of our larger Small model is consistently better than that of Mini across all benchmarks. These results highlight the potential of further scaling dLLMs.

\section{Acknowledgements}
We are grateful to the teams at Artificial Analysis and Copilot Arena for their support in independent third-party evaluation of our models.

\bibliography{refs}
\bibliographystyle{neurips_2024}

\end{document}